\documentclass[conference]{IEEEtran}  % Comment this line out if you need a4paper

\IEEEoverridecommandlockouts 
\usepackage{cite}
\usepackage{multirow}
\usepackage{times}
\usepackage{epsfig}
\usepackage{graphicx}
\usepackage{lipsum}
\usepackage{textcomp}
\usepackage{xcolor}
\usepackage{algorithmic}
\usepackage{amsmath,amssymb,amsfonts}
\usepackage{url}
\usepackage{upgreek}
 \usepackage{amssymb}
\def\BibTeX{{\rm B\kern-.05em{\sc i\kern-.025em b}\kern-.08em
    T\kern-.1667em\lower.7ex\hbox{E}\kern-.125emX}}

\title{\LARGE \bf U-NetPlus: A Modified Encoder-Decoder U-Net Architecture for Semantic and Instance Segmentation of Surgical Instrument}
\author{S. M. Kamrul~Hasan
        and~Cristian A.~Linte,~\IEEEmembership{Senior Member,~IEEE}
\thanks{Research reported in this publication was supported by the National Institute of General Medical Sciences of the National Institutes of Health under Award No. R35GM128877 and by the Office of Advanced Cyber infrastructure of the National Science Foundation under Award No. 1808530.} \vspace{5mm}
\thanks{S. M. Kamrul Hasan is with the Center for Imaging Science, Rochester Institute of Technology, Rochester,
NY USA. E-mail: sh3190@rit.edu.}% <-this % stops a space
\thanks{Cristian A. Linte is with the Biomedical Engineering and Center for Imaging Science, Rochester Institute of Technology, Rochester, NY USA. Email: calbme@rit.edu}}% <-this % stops a space

\begin{document}

\maketitle
% \IEEEpubidadjcol

\begin{abstract}
    Conventional therapy approaches limit surgeons' dexterity control due to limited field-of-view. With the advent of robot-assisted surgery, there has been a paradigm shift in medical technology for minimally invasive surgery. However, it is very challenging to track the position of the surgical instruments in a surgical scene, and accurate detection \& identification of surgical tools is paramount. Deep learning-based semantic segmentation in frames of surgery videos has the potential to facilitate this task. In this work, we modify the U-Net architecture named U-NetPlus, by introducing a pre-trained encoder and re-design the decoder part, by replacing the transposed convolution operation with an upsampling operation based on nearest-neighbor (NN) interpolation. To further improve performance, we also employ a very fast and flexible data augmentation technique. We trained the framework on $8 \times 225$ frame sequences of robotic surgical videos, available through the MICCAI 2017 EndoVis Challenge dataset and tested it on $8 \times 75$ frame and $2 \times 300$ frame videos. Using our U-NetPlus architecture, we report a 90.20\% DICE for binary segmentation, 76.26\% DICE for instrument part segmentation, and 46.07\% for instrument type (i.e., all instruments) segmentation, outperforming the results of previous techniques implemented and tested on these data.
\end{abstract}


%%%%%%%%%%%%%%%%%% BODY TEXT %%%%%%%%%%%%%%%%%%  
\section{Introduction}
 
Minimally invasive surgery has addressed many of the challenges of traditional surgical approaches by significantly reducing risk of infections and shortening hospitalization times, achieving similar outcome to traditional open surgery. There is a new paradigm shift in this field thanks to robot assistance under laparoscopic visualization \cite{rao2018robotic}. To facilitate the manipulation of the laparoscopic surgical instruments while visualizing the endoscopic scene, surgical instrument identification is critical. Nevertheless, this task is challenging, due to the surrounding effects like illumination changes, visual occlusions, and presence of non-class objects. Hence, it is important  to  devise segmentation techniques  that  are  sufficiently  accurate  and  robust  to  ensure accurate tracking of the surgical tools to facilitate therapy via accurate manipulation of the laparoscopic instruments.

Although in recent years semantic segmentation methods applied to  city-scapes, street scenes, and even Landsat image datasets \cite{yang2018denseaspp,he2017mask} have achieved ground-breaking performance by the virtue of deep convolutional neural networks (CNNs), image segmentation in clinical settings still requires more accuracy and precision, with even minimal segmentation errors being unacceptable. In the context of deep learning, Long {\it et al.} \cite{long2015fully} proposed the first fully convolutional network (FCN) for semantic image segmentation, exploiting the capability of Convolutional Neural Networks (CNNs). However, their adoption for use in the medical  domain was initially challenging, due to the limited availability of medical imaging data. These challenges were later circumvented by patch-based training, data augmentation, and transfer learning techniques \cite{Shen:2017}. These newer deep architectures learn to decode low-resolution images produced by VGG16 network into pixel-wise predictions. This network has 13 convolutional layers, with 3 fully connected layers and their weights are typically pre-trained on the large ImageNet object classification dataset \cite{imagenet_cvpr09}, whereas, the decoder network upsamples feature maps from the bottleneck layer. 

However, semantic segmentation is not sufficiently accurate for handling multi-class objects, due to the close presence of objects of the same class in the surgical scene. Therefore, the proposed work is motivated by the need to improve multi-class object segmentation, by leveraging the power of the existing U-Net architecture and augmenting it with new capabilities.

With the advent of U-Net architectures, a wide range of medical imaging tasks have been implemented and produced state-of-the-art results since 2015 \cite{ronneberger2015u}. Recently, Chen \textit{et al.} modified U-Net architecture by introducing sub-pixel layers to improve low-light imaging \cite{chen2018learning}. Following the end-to-end training of a fully-convolutional network, they obtained promising results, with high signal-to-noise-ratio (SNR) and perfect color transformation on their own SID dataset. The authors in \cite{jiang2015quantum, jia2017super} used nearest-neighbor interpolation for image reconstruction and super-resolution. The authors in \cite{odena2016deconvolution} investigated the problem of transposed convolution and provided a solution by  nearest-neighbor interpolation. However, the importance of integrating it into the deep CNN as part of the image upsampling operation was not fully explored so far. There have been a few papers tackling the segmentation and identification of surgical instruments from endoscopic video image, and, even fewer than half a dozen papers tackling this challenge using deep learning. One notable research contribution has been the use of a modified version of FCN-8, yet with no attempts for multi-class segmentation \cite{garcia2017toolnet}. 

Multi-class (both instrument part and type) tool segmentation was first proposed by Shvets \textit{et al.} \cite{shvets2018automatic}, and Pakhomov \textit{et al.} \cite{pakhomov2017deep} and achieved promising results. They modified the classic U-Net model \cite{ronneberger2015u} that relies on the transposed convolution or deconvolution, in a similar, yet opposite fashion to the convolutional layers. As an example, instead of mapping from $4 \times 4$ input pixels to 1 output pixel, they map from 1 input pixel to $4 \times 4$ output pixels. However, its performance is much slower as the filters need additional weights and parameters that also require training. These large number of trainable parameters make the network hard to train end-to-end on a relevant task. Additionally, transposed convolution can easily lead to ``uneven overlap'', characterized by checkerboard-like patterns resulting in artifacts on a variety of scales and colors \cite{odena2016deconvolution}. Redford \textit{et al.} \cite{radford2015unsupervised} and Salimans \textit{et al.} \cite{salimans2016improved}
mentioned the problem of those artifacts and checkerboard patterns generated by the transposed convolution. While it is difficult to entirely remove these limitations and their resulting artifacts, our goal is to, at first,  minimize their occurrence. 

Hence, in the efforts to mitigate these challenges associated with the classic U-Net architecture, in this paper, we present the U-NetPlus model by introducing both VGG-11 and VGG-16 as an encoder with batch-normalized pre-trained weights and nearest-neighbor interpolation as the replacement of the transposed convolution in the decoder layer. This pre-trained encoder \cite{arXiv:1801.05746} speeds up convergence and leads to improved results by circumventing the optimization challenges associated with the target data \cite{he2018rethinking}. Moreover, the nearest-neighbor interpolation used in the decoder section removes the artifacts generated by the transposed convolution. 
%%%%%%%%%%% Figure 1 %%%%%%
%########################
%########################
%########################
%########################
\begin{figure}[ht!]
%\begin{center}

\includegraphics[width=1.0\linewidth]{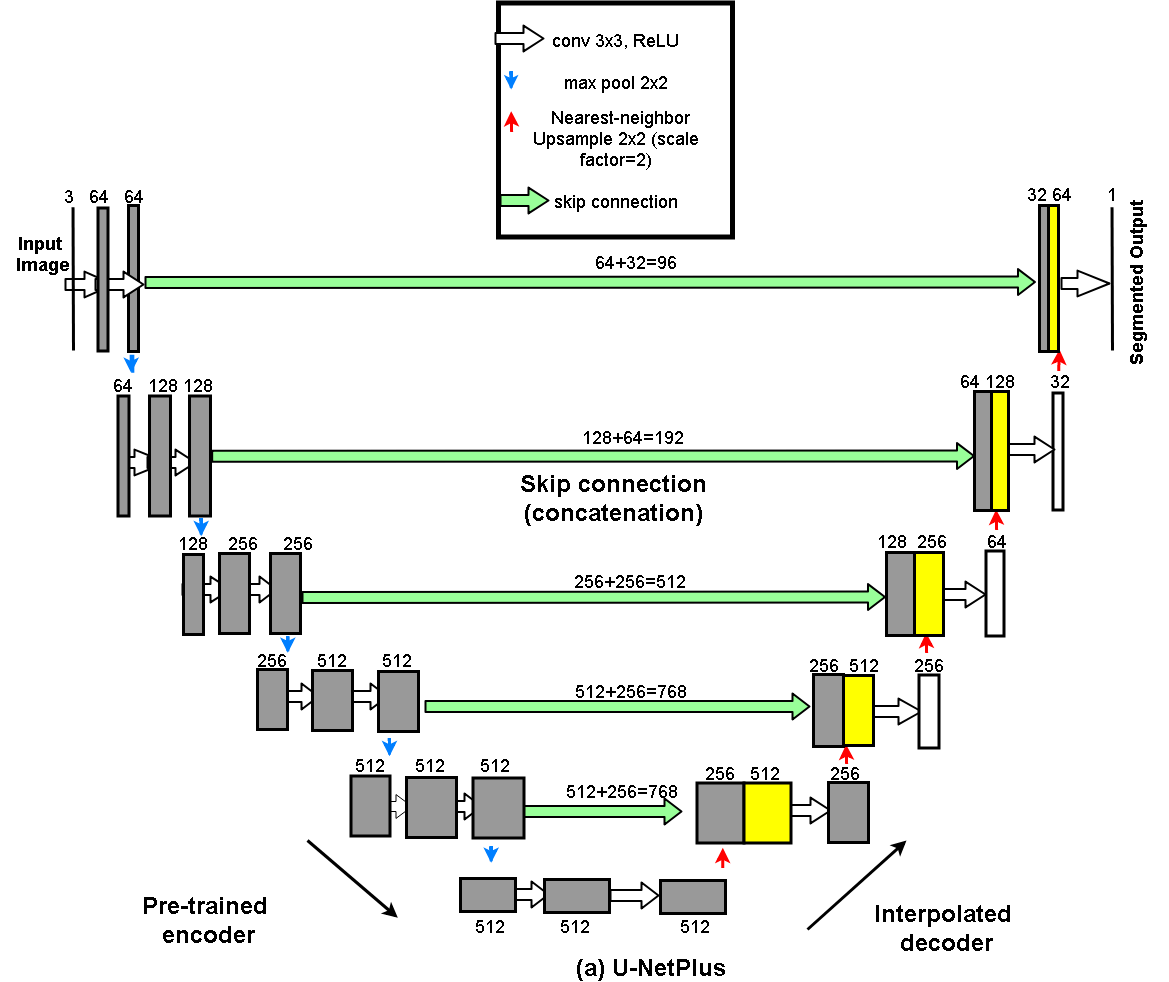}
\includegraphics[width=1.0\linewidth]{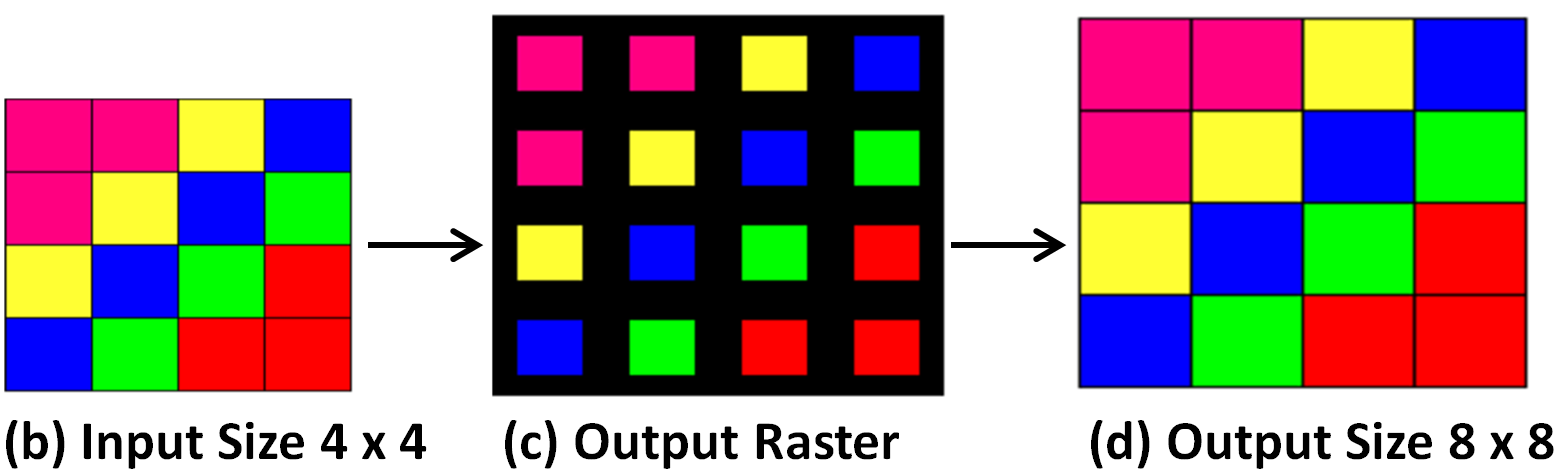}
%\end{center}
   \caption{(a) Modified U-Net with batch-normalized VGG11 as an encoder. Each box corresponds to a multi-channel featuring a map passing through a series of transformations. It consists of both an upsampling and a downsampling path and the height of the box represents the feature map resolution, while the width represents number of channels. Cyan arrows represent the max-pooling operation, whereas light-green arrows represent skip connections that transfer information from the encoder to the decoder. Red upward arrows represent the decoder which consists of nearest-neighbor upsampling with a scale factor of 2 followed by 2 convolution layers and a ReLU activation function; (b)-(d) working principle of nearest-neighbor interpolation where the low-resolution image is resized back to the original image.}
\label{fig:architecture}
\label{fig:NN}
\end{figure}

To test the proposed U-NetPlus network, we implemented some of the recent state-of-the-art architectures for surgical tool segmentation and compared their results to those of the U-NetPlus architecture. From the above mentioned papers, only one seems to have achieved results comparable to ours \cite{arXiv:1801.05746}, but it still suffers from several artifacts, which we have been able to further mitigate some of these artifacts using our proposed method. As such, while this paper leverages some of the existing infrastructure of fully convolutional network, it focuses on demonstrating the adaptation of existing infrastructure to refine its performance for a given task --- in this case the segmentation and identification of surgical instruments from endoscopic images --- rather than proposing a new fully convolutional framework. We believe there exist sufficient tools for segmentation and identification, however the integration and adaptation of these tools for improved performance is key to further improving the outcome of such tools. We demonstrate that the potential use of nearest-neighbor interpolation in the decoder removes artifacts and reduces the number of parameters.

%%%%%%%%%%%%%%%%%% Methodology %%%%%%%%%%%%%%%%%%  

\section{Methodology}
% To address the limitations associated with the classical U-Net architecture, we have focused on improving the architecture by applying a modified U-Net, like TernausNet \cite{arXiv:1801.05746} architecture, for multi-class segmentation purpose.

\subsection{Overview of Proposed Method}
U-NetPlus has an encoder network and a corresponding decoder network, followed by a final pixel-wise segmentation layer, as illustrated in the architecture in Fig. \ref{fig:architecture}. Similar to U-Net, our proposed U-NetPlus works like an auto-encoder with both a downsampling and an upsampling path. To maintain the number of channels the same as in the encoder part, skip connections are added between blocks of the same size in the downsampling and upsampling paths. This allows very precise alignment of the mask to the original image, which is particularly important in medical imaging. Furthermore, skip connections are known to ease network optimization by introducing multiple paths for backpropagation of the gradients, hence, mitigating the vanishing gradient problem.
%\includegraphics[width=7.5cm,height=3.5cm]{{BRATS2017.png}}

% \begin{figure}[htb]%
%     \centering
%     \subfloat{{\includegraphics[width=10.5cm,height=8.5cm]{figures/sub_ter.png}}}
%     \qquad
%     \subfloat{{\includegraphics[width=0.2\linewidth]{figures/NN.png} }}%
%     \caption{(a) Modified U-Net with batch-normalized VGG11 as an encoder. Each box corresponds a multi-channel features map passing through a series of transformations. It consists of both upsampling path and downsampling path and the height of the box represent the feature map resolution, while the widths represent number of channels. Black down-arrows represent the max-pooling operation, whereas gray arrows are for skip connections that transfers information from encoder to decoder. Up-arrows are the decoder which consists of nearest-neighbor upsampling with scale factor of 2 followed by 2 convolution layers and ReLU activation function; (b)-(d) working principle of nearest-neighbor interpolation where the low-resolution image get resized back to original image.}%
%     \label{fig:example}%
% \end{figure}

Generally, weights are initialized randomly to train a network. However, limited training data can introduce overfitting problems, which become very ``expensive'' as far as manually altering the segmentation mask. Therefore, transfer learning can be used to initialize the network weights. But since a surgical instrument is not a class of ImageNet, one way to use transfer learning for a new task is to partially reuse ImageNet feature extractor --- VGG11 or VGG16 as encoder --- and then add a decoder. An improvement has been introduced for the encoder part, where we initiated a pre-trained VGG-11 and VGG-16 architecture with batch-normalization layers that has 11 and 16 sequential layers, respectively. Following this modification, it has been shown the pre-trained model is able to train the network within a very short time and with greater accuracy \cite{kornblith2018better}.

The feature map of VGG11 consists of seven convolutional layers of $3 \times 3$  kernel size  followed by a ReLU activation function. For the reduction of the feature map size, max polling with stride 1 was used. The network starts by producing 64 channels in its first layer, and then continues by doubling the number of channels after each  pooling operation until 512. Weights are copied from the original pre-trained VGG-11 on Imagenet.

The key effect of batch normalization has been investigated in a recent paper \cite{santurkar2018does}. According to this work, batch normalization not only reduces the internal co-variate shift, but also re-parameterizes the underlying gradient optimization problem that makes the training more predictive at a faster convergence. After analyzing the impact of inserting BatchNorm layer, we applied BatchNorm layer after each convolutional layer.

The downsampling path decreases the feature size while increasing the number of feature maps, whereas the upsampling path increases the feature size while decreasing the number of feature maps, eventually leading to a pixel-wise mask. For the upsampling operation, we modified the existing architecture to reconstruct the high-resolution feature maps. Rather than using transposed convolution, we used the nearest-neighbor upsampling layer with a carefully selected stride and kernel size at the beginning of each block followed by two convolution layers and a ReLU function that would increase the spatial dimension in each block by a factor of 2. 

Nearest-neighbor interpolation upsamples the input feature map by superimposing a regular grid onto it. Given $I_i$
be the input grid which is to be sampled, the output grid is produced by a linear transformation $\tau_{\theta}$($I_i$). Therefore, for an upsampling operation, $\tau_{\theta}$ can be defined as:

\begin{equation}
    \begin{pmatrix} p_{i}^o \\ q_{i}^o \end{pmatrix} = \tau_{\theta} (I_{i}) =
    \begin{bmatrix} 
    \theta & 0 \\
    0 & \theta 
    \end{bmatrix}
    \begin{pmatrix} p_{i}^t \\ q_{i}^t \end{pmatrix},  \theta \geq 1,
\end{equation}
where $(p_{i}^o , q_{i}^o)$ $\epsilon$  $I_{i}$ are the original sampling input coordinates, $(p_{i}^t , q_{i}^t)$ are the target coordinates, and $\theta$ upsampling factor. The principle of how nearest-neighbor (NN) interpolation works to enlarge the image size, is shown in the Fig. \ref{fig:NN}. After locating the center pixel of the cell of the output raster dataset on the input raster, the location of the nearest center of the cell on the input raster will be determined and the value of that cell on the output raster will be assigned afterwards. As an example, we demonstrate the upsampling of a $4 \times 4$ image using this approach. The cell centers of the output raster are equidistant. A value is needed to be derived from the input raster for each output cell. Nearest-neighbor interpolation would select those cells centers from the input raster that are closest to that of output raster. The black areas of the middle image can be filled  with the copies of the center pixel. Therefore, this fixed interpolation weights requires no learning for upsampling operation compared to strided or transposed convolution leading to create a more memory efficient upsampling operation. The algorithm is similar to the one proposed and used by the authors of \cite{dong2016accelerating} in their work.
%#############################
%#############################
%############Figure 2 ########
%#############################

\begin{figure}[t]
\begin{center}
    \includegraphics[width=1.0\linewidth]{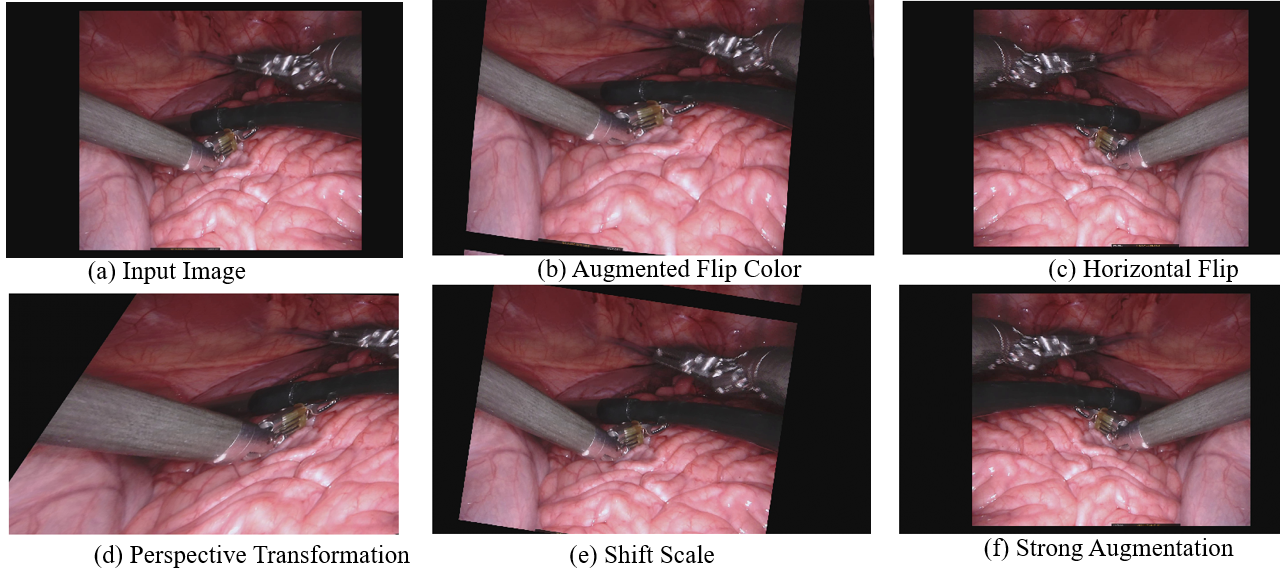}
    \end{center}
      \caption{Example images of applying both affine and elastic transformation in albumentations library for data augmentation.}
    \label{fig:augmented}
\end{figure}

\subsection{Dataset}
 For both training and validation, we used the Robotic instruments dataset from the sub-challenge of \textbf{MICCAI 2017 Endoscopic Vision Challenge} \cite{Endovis:2017}. The training dataset has $8\times 225$ frame sequences with 2 Hz frame rate of high resolution stereo camera images collected from a da Vinci Xi surgical system during laparoscopic cholecystectomy procedures. The frames were re-sampled from 30 Hz video to 2 Hz to avoid any redundancy issues. A stereo camera was used to capture the video sequences that consists of the left and right eye views with resolution of $1920 \times 1080$ in RGB format. In each frame, the articulated parts of the robotic surgical instrument consisting of a rigid shaft, an articulated wrist, and claspers, have been manually labeled by expert clinicians. The test set has $8 \times 75$ frame sequences and $2 \times 300$ frame videos. The challenge is to segment 7 classes such
as prograsp forceps, needle driver, vessel sealer, grasping retractor etc.

\subsection{Data Augmentation}
We augmented the MICCAI 2017 EndoVis Challenge data using the albumentations library that was reported as a fast and flexible implementation for data augmentation in \cite{2018arXiv180906839B}. These libraries include both affine and elastic transformations, and their effects on the image data during augmentation is illustrated in Fig. \ref{fig:augmented}. 

In short, the affine transformation includes scaling, translation, horizontal flip, vertical flip, random brightness and noise addition etc. For the elastic transformation (non-affine), first a random displacement field, $F(R)$ is generated for the horizontal and vertical directions, \ensuremath{\delta x}, and \ensuremath{\delta y} respectively where, \lbrack \ensuremath{\delta x}, \ensuremath{\delta y} \rbrack = \lbrack -1 \ensuremath{\leq} \ensuremath{\delta x}, \ensuremath{\delta y}  \ensuremath{\leq} +1 \rbrack.

These random fields are then convolved with an intermediate value of \ensuremath{\sigma} (in pixels) and the fields are multiplied by a scaling factor \ensuremath{\alpha} that controls intensity. Thus, we obtain the elastically transformed image in which the global shape of the interest is undisturbed, unlike in the affine-transformed image.

\subsection{Implementation Details}
 We implemented our methodology using PyTorch\footnote{https://github.com/pytorch/pytorch}. During the pre-processing step, we cropped the un-wanted black border from each video frame. Images were normalized by subtracting their mean and dividing by their standard deviation (i.e., according to their z-scores). Batch normalization was used before each weighted layer, as it re-parameterizes the underlying gradient optimization problem that helps the training to converge faster \cite{santurkar2018does}. For training, we use the Adam optimizer with a learning rate of 0.00001. We didn't use dropout as it degraded validation performance in our case. All models were trained for 100 epochs. The training set was shuffled before each epoch and the batch size was 4 in our case. All experiments were run on a machine equipped with a NVIDIA GTX 1080 Ti GPU (11GBs of memory).

% The key idea of using DSC and IoU as the performance metrics is that they works well when the foreground pixel is small compared to the background. In our case, tool pixels are small compared to the background pixels.
\subsection{Performance Metrics}
In this work, we used the common Jaccard index --- also referred to as the intersection-over-union (IoU) --- to evaluate segmentation results. It is an overlap index that quantifies the agreement between two segmented image regions: a ground truth segmentation and the predicted segmentation method. Given a vector of ground truth labels $T_{1}$ and a vector of predicted labels $P_{1}$, IoU can be defined as

\begin{equation} \label{eq:1}
   J(T_{1}, P_{1}) = {}{{\frac{|T_{1} \cap P_{1} |}{|T_{1} \cup P_{1}|}}} \\
    = {{\frac{|T_{1} \cap P_{1}|}{|T_{1}| + |P_{1}| - |T_{1} \cap P_{1}|}},}
\end{equation}

Eqn. \ref{eq:1}  can further be clarified. Given a pixel $j$, the label of the pixel $z_j$, and the probability of the same pixel for the predicted class $\hat{z_j}$, Eqn. 1 for $k$ number of dataset
\begin{equation}
 	J=\frac{1}{k}\displaystyle\sum_{j=1}^{k} (\frac{z_j\hat{z_j}}
 	{z_j + \hat{z_j} - z_j \hat{z_j}}),
\end{equation}

We can represent the loss function in a common ground of $log$ scale as this task is a pixel classification problem. So, for a given pixel $j$, the common loss can be defined as the function $H$ for $k$ number of dataset
\begin{equation}
 	H=-\frac{1}{k}\displaystyle\sum_{j=1}^{k} (z_j\log\hat{z_j} + (1- z_j)\log(1-\hat{z_j})),
\end{equation}
From both the Eqn. 1 and Eqn. 2, we can combine and can get a generalized loss 
\begin{equation}
L= H - \log{J}
\end{equation}
Our aim is to minimize the loss function, and, to do so, we can maximize the probabilities for correct pixels to be predicted and maximize the intersection, $J$ between masks and corresponding predictions. 

Another commonly used performance metric is the DICE coefficient. Given the set of all pixels in the image, set of foreground pixels by automated segmentation $S_{1}^a$, and the set of pixels for ground truth $S_{1}^g$, DICE score can be compared with [$S_{1}^a$, $S_{1}^g$]\ensuremath{\subseteq} \ensuremath{\Omega}, when a vector of ground truth labels $T_{1}$ and a vector of predicted labels $P_{1}$,

\begin{equation}
D(T_{1}, P_{1}) = {}{{\frac{2|T_{1} \cap P_{1} |}{|T_{1}| + |P_{1}|}}}
\end{equation}
DICE score will measure the similarity between two sets, $T_{1}$ and $P_{1}$ and $|T_{1}|$ denotes the cardinality of the set $T_{1}$ with the range of D($T_{1}$,$P_{1}$) \ensuremath{\epsilon} [0,1].

%Figure 3%
\begin{figure*}[ht!]
\begin{center}
\includegraphics[width=1.0\linewidth]{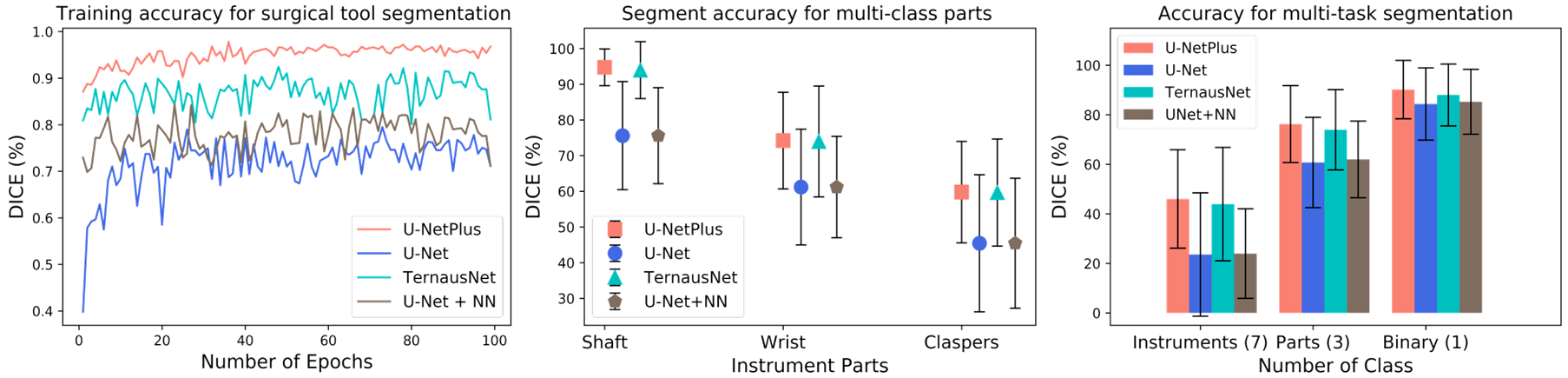}

\end{center}
  \caption{ Quantitative comparison of (a) training accuracy (left), (b) multi-class (class=3)
instrument parts (middle) (c) multi-task segmentation accuracy (right). }
\label{fig:Train}
\label{fig:Accuracy}
\end{figure*}

%%%%%%%%%%%%%%%%%% Table 1 %%%%%%%%%%%%%%%%%%  
%%%%%%%%%%%%%%%%%% Table 1 %%%%%%%%%%%%%%%%%%  
%%%%%%%%%%%%%%%%%% Table 1 %%%%%%%%%%%%%%%%%%  
%\begin{scriptsize}
\begin{center}
\begin{table*}[ht]
\vspace{-2mm}
\centering
\label{table:test}
\caption{\label{tab:somelabel} Quantitative evaluation of the segmentation results. Mean and (standard deviation) values are reported for IoU(\%) and DICE coefficient(\%) from all networks against the provided reference segmentation. The statistical significance of the results for U-Net + NN and U-NetPlus model compared against the baseline model (U-Net and TernasuNet) are represented by $*$ and $**$ for p-values 0.1 and 0.05, respectively. U-Net has been compared with U-Net+NN, TernausNet has been compared with U-NetPlus. The best performance metric (IoU and DICE) in each category (Binary, Instrument Part and Instrument Type Segmentation) is indicated in \textbf{bold} text.}
\begin{tabular}{|c||cc|cc|cc|}
\hline
%\toprule
    \textbf{Network} & \multicolumn{2}{c}{\textbf{Binary Segmentation }} & \multicolumn{2}{c}{\textbf{Instrument Part}} & \multicolumn{2}{c|}{\textbf{Instrument Type}} \\
     & & & & & &\\
  
%  \textbf{} & \multicolumn{2}{c}{\head{\textbf{}}} & \multicolumn{2}{c}{\head{\textbf{}}} & \multicolumn{2}{c|}{\head{\textbf{}}}\\
\hline
\textbf{Metric} & \textbf{IoU} & \textbf{DICE} & \textbf{IoU} & \textbf{DICE} & \textbf{IoU} & \textbf{DICE}\\
\hline \hline
	ToolNetH \cite{garcia2017toolnet} & 74.4  & 82.2 & -  & - & -  & -\\
	\hline
	ToolNetMS \cite{garcia2017toolnet} & 72.5  & 80.4 & -  & - & -  & - \\
	\hline
	FCN-8s \cite{garcia2017toolnet}& 70.9  & 78.8& -  & -& -  & - \\
	\hline
	CSL \cite{laina2017concurrent} & -  & 88.9 & -  & 87.70 (Shaft) & -  & - \\
	\hline
	U-Net\cite{ronneberger2015u} & 75.44  & 84.37& 48.41&	60.75 &	15.80 &	23.59 \\
	 & (18.18)  & (14.58) & (17.59) & (18.21) &	(15.06) &	(19.87) \\
    \hline
	\textbf{U-Net + NN} & \textbf{77.05**}  & \textbf{85.26*} & \textbf{49.39*} &	\textbf{61.98*} &	\textbf{16.72*} &	\textbf{23.97} \\
	 & (15.71)  & (13.08) & (15.18) & (15.47) &	(13.45) &	(18.08) \\
	 \hline
	TernausNet \cite{shvets2018automatic} & 83.60  & 90.01  & 65.50  & 75.97 & 33.78  & 44.95 \\
	 & (15.83)  & (12.50) & (17.22) & (16.21) &	(19.16) &	(22.89) \\
	\hline
	\textbf{U-NetPlus-VGG-11} & 81.32  & 88.27 & 62.51  & 74.57 & \textbf{34.84*}  & \textbf{46.07**} \\
	 & (16.76)  & (13.52) & (18.87) & (16.51) &	(14.26) &	(16.16)  \\
	\hline
	\textbf{U-NetPlus-VGG-16} & \textbf{83.75}  & \textbf{90.20*} & \textbf{65.75}  & \textbf{76.26*} &  34.19  & 45.32 \\
		  & (13.36)  & (11.77) & (14.74) & (13.54) &	(15.06) &	(17.86) \\
		   &   &  &  & \textbf{94.75(Shaft)} &  &  \\
	\hline
\end{tabular}
\end{table*}
\end{center}
%\end{scriptsize}

%%%%%%%%%%%%%%%%%% Attention %%%%%%%%%%%%%%%%%%  
%%%%%%%%%%%%%%%%%% Attention %%%%%%%%%%%%%%%%%%  
%%%%%%%%%%%%%%%%%% Attention %%%%%%%%%%%%%%%%%%  
\section{Results}
\subsection{Quantitative Results} 
To illustrate the potential improvement in segmentation performance by using the nearest-neighbor interpolation (i.e., fixed upsampling) in the decoder, we conducted a pair comparison between the segmentation results obtained using the classical U-Net architecture, U-Net + NN, TernausNet, and U-NetPlus (our proposed method).

Training accuracy for binary segmentation is shown in Fig. \ref{fig:Accuracy} for 100 epochs. We compare our proposed architecture with three other models: U-Net, U-Net+NN, TernausNet. We can observe from this figure that after adding nearest-neighbor (NN) in the decoder of U-Net, the training accuracy of the classical U-Net framework (shown in blue) featuring the transposed convolution in the decodes, improves. Furthermore, the training of our proposed method (U-NetPlus) also converges faster and yields better training accuracy compared to TernausNet (shown in cyan). Hence, this graph alone illustrates the benefit of the nearest-neighbor interpolation on the segmentation performance.

%%%%%%%%%%%%%%%%%% Result %%%%%%%%%%%%%%%%%%  
%%%%%%%%%%%%%%%%%% Result %%%%%%%%%%%%%%%%%%  
%%%%%%%%%%%%%%%%%% Result %%%%%%%%%%%%%%%%%%  

The model was tested on the MICCAI 2017 EndoVis dataset. Table \ref{tab:somelabel} summarizes the performance of our proposed U-NetPlus framework in the context of several state-of-the art multi-task segmentation techniques. The table clearly indicates the improvement in segmentation following the addition of nearest-neighbor interpolation in the decoder step across all frameworks --- U-Net and TernausNet. Moreover, our model had been compared with four different structures other than U-Net and TernausNet --- ToolNetH, ToolNetMS, FCN-8s, and CSL. The last one (CSL) was the first approach to multi-class surgical instrument segmentation. But, they used only two instrument classes (shaft and claspers) and omit wrist class which we introduced in our approach and the overall accuracy that we obtained was significantly higher than CSL approach.

We conducted a paired statistical test to compare the segmentation performance of each of these methods (U-Net, U-Net+NN, TernausNet, U-NetPlus) in terms of the IoU and DICE metric. As illustrated, our proposed U-NetPlus architecture yielded a statistically significant\footnote{Wilcoxon signed-rank test is performed for statistical significance testing} 11.01\% improvement (p $<$ 0.05) in IoU and 6.91\%  DICE (p $<$ 0.05) over the classical U-Net framework; a statistically significant 8.0\% improvement (p $<$ 0.05) in IoU and 5.79\% DICE (p $<$ 0.05) over the U-Net + NN framework; a statistically significant 0.18\% improvement in IoU and 0.21\% DICE (p $<$ 0.1) over the state-of-the-art TernausNet framework \cite{shvets2018automatic}.

Multi-class instrument segmentation was performed by labeling each instrument pixel with the corresponding index given in the training set. This application consisted of three classes: shaft, wrist, and claspers. The multi-class segmentation using our proposed U-NetPlus framework yielded a mean 65.75\% IoU and 76.26\% DICE. The accuracy and precision of the U-NetPlus architecture relative to the other three frameworks is illustrated in Fig. \ref{fig:Accuracy}. As shown, the U-NetPlus framework outperforms the currently deemed best-in-class TernausNet framework.

The instrument type was segmented by labelling each instrument pixel with the corresponding instrument type, according to the training set, and all background pixels were labeled as 0. In the case of instrument type segmentation (for class = 7), U-NetPlus-VGG-11 encoder worked better than the U-NetPlus-VGG-16. Our results for instrument type segmentation can be further refined. 

\begin{figure*}[ht]
\includegraphics[width=1.0\linewidth]{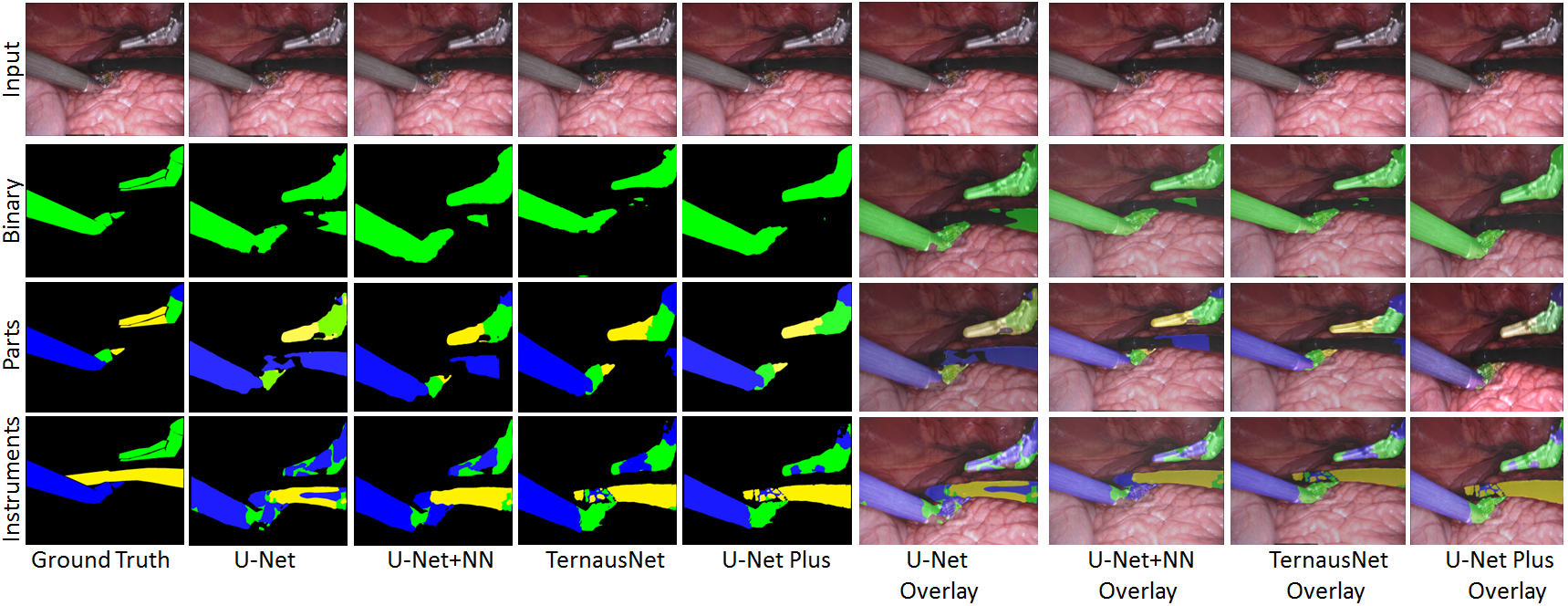}
  \caption{ Qualitative comparison of binary segmentation, instrument part and instrument type segmentation result and their overlay onto the native endoscopic images of the MICCAI 2017 EndoVis video dataset yielded by four different frameworks: U-Net, U-Net+NN, TernausNet, and U-NetPlus.}
\label{fig:UNetPlus}
\end{figure*}

\begin{figure}[t]
    \includegraphics[width=1.0\linewidth]{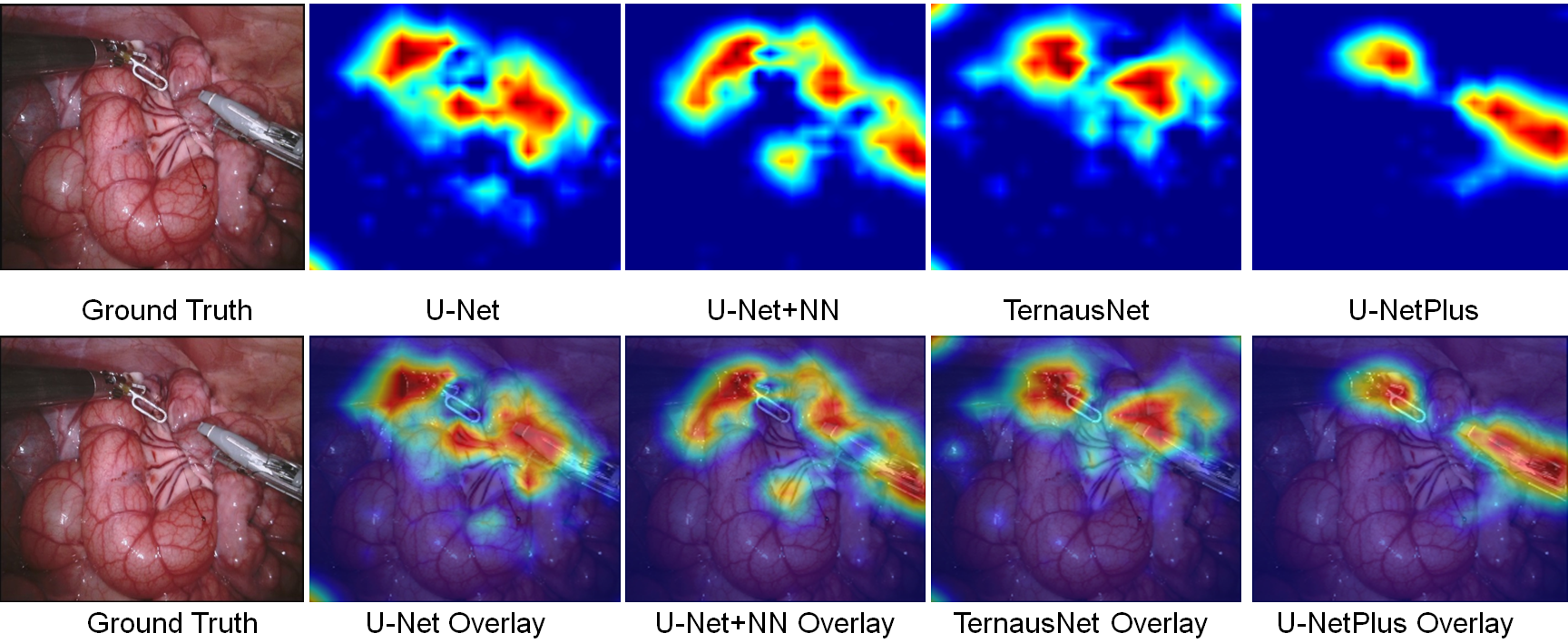}
      \caption{Attention results: U-NetPlus ``looks'' at a focused target region, whereas U-Net, U-Net+NN and TernausNet appear less ``focused'', leading to less accurate segmentation.}
    \label{fig:attention}
\end{figure}

\subsection{Qualitative Results} 
The qualitative performance of our model both for a binary and multi-class segmentation is shown in Fig. \ref{fig:UNetPlus}. The second row of the figure shows that for the binary segmentation, the classical U-Net shows a portion of instrument which was not present in the binary mask of our ground truth data (second row and second column). U-netPlus shows the best performance for binary segmentation (i.e. it can clearly segment out the instruments from the background), whereas TernausNet still shows un-wanted regions in the segmentation output. For the instrument parts segmentation, U-Net still segments the un-wanted instrument (blue), whereas U-NetPlus can segment the 3 classes (blue:shaft, green:wrist, yellow:claspers) near perfectly compared to TernausNet. For the instrument type segmentation, we can clearly observe that U-Net can not differentiate between the blue and the green class, whereas U-NetPlus can differentiate these classes more accurately than TernausNet. Both the binary and multi-class segmented output have been overlayed onto the original image (sixth, seventh, eighth, and ninth column). The figure has a clear indication of qualitative improvement of U-NetPlus over U-Net, U-Net+NN and TernausNet.

\subsection{Attention Study}
To further analyze the improvement in segmentation performance, we performed an attention analysis to visualize where our proposed algorithm ``looks'' in an image by using a novel image saliency technique \cite{fong2017interpretable} that learns the mask of an image by suppressing the softmax probability of its target class. Fig. \ref{fig:attention} shows that using this class activation mapping, our approach (U-NetPlus) localizes the wrist and claspers of the bipolar forceps near perfectly compared to the classical U-Net, U-Net+NN and TernausNet frameworks. U-Net does not perform well compared to U-Net+NN where nearest-neighbor upsampling is added to the decoder path. TernausNet performs well than U-Net+NN due to the use of pre-trained VGG network in encoder. Pre-trained network performs well in this segmentation due to the limited dataset. Fig. \ref{fig:attention} shows the heatmap 
image which has been overlayed onto it's original image.
Therefore, the skillful integration and combination of pre-trained encoder and nearest-neighbor interpolation as a fixed upsampling technique yields higher overall performance.

\section{Discussion and Conclusion}
In this paper, we proposed a modified U-Net model for surgical tool segmentation. To improve  robustness  beyond  that  of  the U-Net  framework, we used the pre-trained model as the encoder with batch-normalization, which converges much faster in comparison to the non-pre-trained network. In the decoder part, we substituted  the  deconvolution layer  with  an upsampling  layer  that  uses nearest-neighbor interpolation followed by  two  convolution  layers. Moreover, we used a fast and effective data augmentation technique to avoid the overfitting problem. We evaluated its performance on the MICCAI 2017 EndoVis Challenge dataset. We also visualized the output of our proposed method both as stand-alone surgical instrument segmentation, as well as overlays onto the native endoscopic images. Apart from that, we also conducted an ``attention'' study to determine where our proposed algorithm ``looks'' in an image. 

Our proposed model with batch-normalized U-NetPlus-VGG-16 outperforms existing methods in terms of both Jaccard and DICE, achieving 90.20\% DICE for binary class segmentation and 76.26\% for parts segmentation, both of which showed at least 0.21\% improvement over the current methods and more than 6\% improvement over the traditional U-Net architecture. Nevertheless, U-NetPlus-VGG-16's performance with regards to identifying the instrument type was inferior to that of U-NetPlus-VGG-11, which was slightly superior to the other disseminated techniques rendering this paper as a first demonstration of a modified version of U-Net Decoder via nearest-neighbor interpolation to remove artifacts induced by the transposed convolution being used for surgical instrument segmentation application and showing improved performance over the TernausNet framework.

\bibliographystyle{IEEEtran}
\bibliography{main}

\end{document}